\documentclass[conference,compsoc]{IEEEtran}

\ifCLASSOPTIONcompsoc
  \usepackage[nocompress]{cite}
\else
  \usepackage{cite}
\fi

\ifCLASSINFOpdf
   \usepackage[pdftex]{graphicx}
   \DeclareGraphicsExtensions{.pdf,.jpeg,.png}
\else
\fi
\usepackage{url}
\usepackage{hyperref}
\usepackage[dvipsnames]{xcolor}
\usepackage{booktabs,tabulary}
\usepackage{xcolor}
\usepackage{tcolorbox}

\tcbset{colback=pink}

\definecolor{light-gray}{gray}{0.4}
\definecolor{dark-gray}{gray}{0.1}

\begin{document}

\title{Helping users discover perspectives: Enhancing opinion mining with joint topic models}

\author{\IEEEauthorblockN{Tim Draws}
\IEEEauthorblockA{TU Delft\\
Email: t.a.draws@tudelft.nl}
\and
\IEEEauthorblockN{Jody Liu}
\IEEEauthorblockA{TU Delft\\
Email: jliu13193@gmail.com}
\and
\IEEEauthorblockN{Nava Tintarev}
\IEEEauthorblockA{TU Delft\\
Email: n.tintarev@tudelft.nl}}

\maketitle

\begin{tcolorbox}
This is an unedited manuscript accepted for presentation at the SENTIRE workshop at the 2020 International Conference on Data Mining (ICDM). The published version of the paper is available at \url{https://doi.org/10.1109/ICDMW51313.2020.00013}.
\end{tcolorbox}

\begin{abstract}
Support or opposition concerning a debated claim such as \emph{abortion should be legal} can have different underlying reasons, which we call \textit{perspectives}. This paper explores how opinion mining can be enhanced with joint topic modeling, to identify distinct perspectives within the topic, providing an informative overview from unstructured text. We evaluate four joint topic models (TAM, JST, VODUM, and LAM) in a user study assessing human understandability of the extracted perspectives. Based on the results, we conclude that joint topic models such as \textbf{TAM} can discover perspectives that align with human judgments. Moreover, our results suggest that users are not influenced by their pre-existing stance on the topic of abortion when interpreting the output of topic models.
\end{abstract}

\IEEEpeerreviewmaketitle

\section{Introduction}

Opinion mining has been used extensively in contexts where it is relevant to extract sentiment, or other richer opinions from unstructured text (e.g., in customer reviews or fora \cite{liu2012survey}). In the context of debated topics, however, it may additionally be important to extract not only a sentiment, or sentiment concerning a single aspect, but also which \textit{sets of aspects} of the debate this sentiment applies to. In other words, to distinguish between different \textit{perspectives} within all stances or sentiments.

For example, a commonly debated claim is \emph{abortion should be legal}. To form an opinion concerning this claim primarily means to take a \emph{stance} (i.e., supporting or opposing the legalization of abortion). However, the same stance can be supported by different underlying reasons~\cite{detectingperspectives}, which we call \emph{perspectives}. For example, someone supporting the legalization of abortion could take the perspective that: \emph{``Reproductive  choice  empowers  women  by  giving  them  control over their own bodies.''}; or instead that: \emph{``Personhood begins after a fetus becomes `viable' (able to survive
outside the womb) or after birth, not at conception.''}

What makes \emph{perspective discovery} (i.e., automatically distilling perspectives from text) challenging is its unstructured nature. Previous research related to debated topics has largely focused on the task of stance classification (i.e., predicting whether a document supports or opposes a given claim), which can be performed in a supervised fashion. In contrast, supervised learning is not applicable to perspective discovery, as pre-defined labels (i.e., the different perspectives on a given topic) usually do not exist.

A family of unsupervised methods that could potentially perform perspective discovery are \textit{topic models}. Topic models aim to find hidden patterns in unstructured corpora of textual documents. Part of the output of a topic model is a pre-defined number of probability distributions (i.e., topics) over all words in the corpus it has been applied to. In practice, topics can then be described by selecting a number of words (e.g., 10) based on the highest probability density in each topic distribution. When applying a topic model on a corpus of opinionated documents, these topics could be seen as perspectives. An example of an output perspective related to abortion could be \textit{\{woman, choice, body, fetus, control, pregnant, birth, baby, foetus, sex\}}. Especially promising in this respect are so-called \emph{joint topic models}, which add additional components (e.g., some form of sentiment analysis) to the classical topic modeling approach. 

Several joint topic models have been developed to serve purposes such as distinguishing ``negative'' or ``opposing'' topics from ``positive'' or ``supporting'' topics (see Section \ref{section:relatedwork_topicmodels} for this related work). Although not all of these models explicitly aim for perspective discovery, their ability to compute topics informed by constructs such as sentiment makes them promising candidates for this task. Something that has -- to the best of our knowledge -- not been evaluated yet is whether joint topic models can perform \emph{human-understandable} perspective discovery. That is, can joint topic models distill perspectives that people can identify?

To study this, we created a data set from debate forum entries on the topic \emph{abortion} and collected perspective annotations for it. We applied several different topic models to this data set. In a user study, we then evaluated whether these topic models are effective in helping people to identify perspectives that exist in the data. We find that a joint topic model such as the \emph{Topic-Aspect Model} (TAM) \cite{twodimensional} can help users distill perspectives from text. Our results furthermore contain no evidence for a tendency of users to interpret topic model output in line with their personal pre-existing attitude. In sum, we make the following contributions:

\begin{itemize}
    \item We publish an openly available, perspective-annotated data set containing 2454 online debate forum entries related to the topic \emph{abortion}.
    \item We present a user study to answer the following two research questions: \textbf{RQ1.} Can joint topic models support users in discovering perspectives in a corpus of opinionated documents?;
\textbf{RQ2.} Do users interpret the output of joint topic models in line with their personal pre-existing stance?
    
\end{itemize}

\noindent All material related to this research (e.g., annotated data set, code, and results) is openly available.\footnote{\url{https://osf.io/uns63/}}

\section{Related work}

Opinion mining has been used extensively to extract sentiment (e.g., positive, negative or neutral) from opinionated text~\cite{liu2012survey}. However, existing methods are limited in terms of extracting perspectives. To better understand this research gap, we first provide an overview of relevant sentiment analysis methods. Next, we relate this work to unstructured text on controversial topics by considering stances in relation to a claim. We finally discuss how topic models can be used to extract perspectives or different aspects that underlie a stance. We conclude by describing how these topic models are enhanced with additional information such as sentiment -- motivating our approach to evaluate them for perspective extraction.

\subsection{Sentiment analysis}
Sentiment analysis (also referred to as \emph{opinion mining}) is the task of deriving sentiment (i.e., a feeling or mood) from text~\cite{liu2020sentiment}. Existing methods for sentiment analysis usually involve learning syntactical structures from text in some supervised fashion. Using subjectivity lexicons, such techniques compute the overall sentiment of a sentence or complete document~\cite{lexicon_taboada, yue2019survey}. For instance, sentiment analysis has been applied to analyze online product reviews~\cite{liu2020sentiment, yue2019survey}.

Big advancements in such techniques in recent years have led researchers to explore applications that aim for deeper levels of text comprehension. One of these sub-fields of sentiment analysis is \emph{stance classification}.

\subsection{Stance classification and aspect mining}
Stance classification is the task of deriving sentiment (i.e., a favorability) towards a specific claim from text~\cite{wang2019survey}. This implies that not just sentiment, but also the \emph{direction} of sentiment needs to be extracted. What makes stance classification challenging is that users may describe their stance in both negative and positive ways. For example, both the following statements imply the same stance, but with different sentimental phrasing: \textit{``I disagree with the terrible idea of legalizing abortion''} and \textit{``It is a very good idea that abortion stays illegal''}. Mere sentiment analysis is thus insufficient for stance classification.

Previously developed techniques for stance classification largely follow supervised approaches~\cite{teja2019controversy}. They are often applied to controversial debates because related opinions are usually either of strong supporting or strong opposing nature. Especially the introduction of \emph{SemEval 2016} (i.e., to classify stance from tweets) has sparked the proposal of several approaches for stance classification~\cite{detectingperspectives,mohammad-etal-2016-semeval}.

Different features have been considered to classify stance. Here, sentiment lexicons and n-grams are among the most used resources, whereas features such as negation, part-of-speech (POS) tags, or punctuation have had a smaller impact~\cite{rosenthal2015semeval}. Rule-based approaches have also been proposed, where syntactical dependency structures or punctuation marks are identified to extract a stance~\cite{argumentationmining}.

Stance classification allows for deeper text comprehension in controversial debates than classical sentiment analysis. To truly understand controversial debates, however, distilling the underlying reasons (i.e., perspectives) behind the different stances is essential. A class of methods that allows for more fine-grained opinion analysis is known as \emph{aspect-based opinion mining} or \emph{aspect mining}~\cite{aspectzhang}. Aspect mining is used to understand the \emph{aspects} or \emph{features} a sentiment is directed at, which could reflect perspectives in the context of controversial topics. Most distributional approaches that have been proposed for aspect extraction (e.g., based on \emph{language rules} and \emph{Hidden Markov Models}) express aspects as single words. \emph{Topic models}, on the other hand, formulate aspects by grouping similar or related words together. This could allow for better descriptions of perspectives compared to other approaches.

\subsection{Topic models}
\label{section:relatedwork_topicmodels}
Topic models are a family of unsupervised models that aim to discover hidden structures in corpora of text. By analyzing word co-occurrences across all documents in a corpus, these models create a previously specified number of \emph{topics}. Each topic is a probability distribution over all words in the corpus. The probability density indicates how ``typical'' a given word is for the topic at hand. This way, topics can be described by their top-n highest-density words.\footnote{Similarly, topic models also output per-document probability distributions over topics to indicate how ``present'' each topic is in a given document.} Arguably the most commonly used topic model today is \emph{Latent Dirichlet Allocation} (LDA)~\cite{ldapaper}.

Joint topic models are a group of models that extend topic modeling (e.g., LDA) by adding components for more informative content extraction from text. For example, several joint topic models within opinion mining have proposed additional distributions or sentiment analysis features on top of LDA to extract more specific aspects. They include the \emph{Topic-Aspect Model} (TAM)~\cite{twodimensional}, the \emph{Joint-Sentiment Topic model} (JST)~\cite{jstmodel}, the \emph{Viewpoint-Opinion Discovery Unified Model} (VODUM)~\cite{vodum}, and the \emph{Latent Argument Model} (LAM) \cite{detectingperspectives}.

Most joint topic models have not specifically been developed for the task of perspective discovery. However, their unsupervised nature and interpretable model output make all joint topic models mentioned above potential candidates in this respect. We compare the various joint topic models for the controversial topic of abortion to evaluate how well existing methods help people discover perspectives from corpora of text.

\begin{table}
\caption{Abortion perspectives in the final data set. Perspectives colored light gray support abortion, whereas perspectives colored dark gray oppose it.}
    \label{tab:6perspectives}
    \centering
    \begin{tabular}{p{0.025\linewidth}p{0.875\linewidth}}
    \toprule
        \textcolor{light-gray}{$p_1$} & \textcolor{light-gray}{Reproductive choice empowers women by giving them control over their own bodies.}\\
        \cmidrule[0.4pt]{1-2}
        \textcolor{light-gray}{$p_2$} & \textcolor{light-gray}{Personhood begins after a fetus becomes 'viable' (able to survive outside the womb) or after birth, not at conception.}\\
        \cmidrule[0.4pt]{1-2}
        \textcolor{light-gray}{$p_3$} & \textcolor{light-gray}{A baby should not come into the world unwanted.}\\
    	\cmidrule[0.4pt]{1-2}
        \textcolor{dark-gray}{$p_4$} & \textcolor{dark-gray}{Abortion is murder, because unborn babies are human beings with a right to life.}\\
    	\cmidrule[0.4pt]{1-2}
        \textcolor{dark-gray}{$p_5$} & \textcolor{dark-gray}{Abortion is the killing of a human being, which defies the word of God.}\\
    	\cmidrule[0.4pt]{1-2}
        \textcolor{dark-gray}{$p_6$} & \textcolor{dark-gray}{If women become pregnant, they should accept the responsibility that comes with producing a child.}\\
    \bottomrule
    \end{tabular}
\end{table}
 
\section{Data}
\label{data}

For this study, we created a perspective-annotated data set consisting of debate forum entries on the topic \emph{abortion}. The data set is openly available.\footnote{ \url{https://osf.io/uns63/}}

\subsection{Creating an annotated data set}
We retrieved a total of $2934$ opinionated documents on the topic \emph{abortion} from an online debate platform.\footnote{\url{https://debate.org}, retrieved May 2020} On this platform, users can participate in openly held debates by posting their opinions in either the supporting or opposing category. 

Each document in our data set was assessed by a human annotator to (1) ensure that all documents are written in English, (2) remove ambiguous documents (such as spam and unclear stance position), and (3) assign a \emph{perspective label} to each document. These perspective labels were taken from the website \emph{ProCon}.\footnote{\url{https://abortion.procon.org}, retrieved May 2020} \emph{ProCon} provides a list of 31 perspectives that exist in the abortion debate (i.e., categorized into \emph{Pro} and \emph{Con}). In the annotation process, it became clear that two perspectives listed at \emph{ProCon} (i.e., \emph{Con 1} and \emph{Con 2}) were difficult to distinguish. We therefore merged these two perspectives into one.\footnote{We formulated this merged perspective as \emph{Abortion is murder, because unborn babies are human beings with a right to life.} (see Table \ref{tab:6perspectives}).}

We controlled the annotation quality by having a randomly selected 10\% documents annotated by another, independent annotator. The results of this quality control suggested that the main annotator was reliable (Krippendorff's $\alpha$ = $0.81$).\footnote{For the annotation reliability metric Krippendorff's $\alpha$, a score of $0.8$ or higher is desired~\cite{measuring_zapf}.}

\subsection{Curating a balanced data set}

For our user study, we aimed to curate a data set that is balanced in terms of stances as well as perspectives. To create this final data set, we picked documents from the raw annotated data to include (1) an equal amount of supporting as well as opposing documents, and (2) an equal amount of documents across six selected perspectives. We selected these six perspectives (i.e., three supporting and three opposing the legalization of abortion; see Table~\ref{tab:6perspectives}) because they were the most commonly occurring perspectives in the data.

We created the final data set by randomly picking $100$ documents from each of the six perspectives listed above. Here we only considered documents that had \emph{uniquely} been annotated with the perspective at hand; thus excluding documents that expressed several different perspectives at once. This resulted in a corpus of $600$ documents that was balanced in terms of stances and perspectives.

\subsection{Preprocessing}
To prepare the final data set for topic modeling, we applied several pre-processing steps. First, we removed any contractions, punctuation, and digits. Second, we lowercased the text and removed stop words. Third, we applied a spelling checker and performed lemmatization. Fourth, we applied antonyms, removed non-sentiment words that do not appear in the subjectivity lexicon \emph{SentiWordNet}~\cite{sentiwordnet} and added bigrams and trigrams.

\section{Method}\label{section:method}

We applied six different models (i.e., four joint topic models and two baseline models) to the data set containing 600 perspective-annotated documents (see Section \ref{data}) and showed parts of the output to participants in a user study. Using sets of keywords, participants had to identify the six correct perspectives that are present in the data. Specifically, participants saw the top ten keywords for each of the six topics that the model at hand had computed.\footnote{It is common practice to represent the output of topic models by the top ten keywords. Accordingly, for our study, we decided that ten words should be enough for participants to understand what the topic is about, but at the same time not too much so that participants are not overwhelmed.}

\subsection{Models}\label{sec:methodModels}

We evaluated four different joint topic models in terms of their ability to help users discover perspectives in corpora of opinionated documents. These joint topic models were TAM, JST, VODUM, and LAM. Each of them performs LDA and adds an additional component where tokens are grouped in a particular way (see Table \ref{tbl:topicmodels}).

To compare the joint topic models to a baseline, we evaluated two additional models (see Table \ref{tbl:topicmodels}). First, we added a regular topic model (i.e., LDA) to test the impact of the components that the joint topic models add on top of LDA. Second, we created a model whose output merely \emph{resembled} that of a topic model by randomly distributing the top 60 words in the corpus (according to \emph{term frequency-inverse document frequency}; TF-IDF) over 6 sets. The purpose of this TF-IDF model was to create a ``control condition'' in which the presented output consists of incoherent groups of words that can still vaguely be associated with the topic \emph{abortion}.

Aside from the TF-IDF model, all models were computed using the original approach and code proposed by their respective authors. In terms of their core topic modeling functionality, each model used similar hyperparameter values to those with which topic models are typically configured~\cite{griffiths2004finding,qiang2020short}. The hyperparameter values were: $1000$ iterations, $\beta = 0.01$, number of topics $k = 6$ (i.e., to reflect six different perspectives), and $\alpha = 50 / k$.

\begin{table}
\caption{Models used in the user study.}
    \label{tbl:topicmodels}
    \centering
    \begin{tabular}{p{0.11\linewidth}p{0.5\linewidth}>{\raggedright\arraybackslash}p{0.24\linewidth}}
        \toprule
        \textbf{Model} & \textbf{Description} & \textbf{Implementation}\\
        \cmidrule[0.4pt]{1-3}
        TF-IDF & A baseline model created by randomly distributing generally important words from the corpus over six groups. & \texttt{Sklearn}~\cite{scikit-learn}\\
        \cmidrule[0.4pt]{1-3}
        LDA & A baseline topic model that computes bag-of-words topics to describe themes in text. & Blei, Ng, \& Jordan \cite{ldapaper}; \texttt{Gensim}~\cite{rehurek_lrec}\\
        \cmidrule[0.4pt]{1-3}
        TAM & Joint topic model that performs LDA and adds additional distributions and processes to group tokens into \emph{background}, \emph{topic-specific}, and \emph{perspective-specific} tokens. & Paul \& Girju \cite{twodimensional}\\
        \cmidrule[0.4pt]{1-3}
        JST & Joint topic model that performs LDA and groups tokens according to a subjectivity lexicon. & Lin, \& He~\cite{jstmodel}\\
        \cmidrule[0.4pt]{1-3}
        VODUM & Joint topic model that performs LDA and groups tokens according to POS-tags. & Thonet, Cabanac, Boughanem, \& Pinel-Sauvagnat~\cite{vodum}\\
        \cmidrule[0.4pt]{1-3}
        LAM & Joint topic model that performs LDA and groups tokens according to a subjectivity lexicon and POS-tags. & Vilares \& He~\cite{detectingperspectives}\\
        \bottomrule
    \end{tabular}
\end{table}

\subsection{Operationalization}\label{section:operationalization}

To compare the models introduced above and investigate the research questions \textbf{RQ1} and \textbf{RQ2}, we conducted an online between-subjects user study. We measured the following variables:


\subsubsection*{\textbf{Independent variable}}
\begin{itemize}
    \item \emph{Model}. Each participant saw the output of one of six different models that they had randomly been assigned to (see Table \ref{tbl:topicmodels} for a model overview).
\end{itemize}

\subsubsection*{\textbf{Dependent variables}}
\begin{itemize}
    \item \emph{Number of correct perspectives found} (\textit{nCor}). This variable measured how many of the six perspectives that truly exist in the corpus were found by participants based on the model output they saw. It could take on seven different values (i.e., integers ranging from 0 to 6).
    \item \emph{Number of opposing perspectives selected} (\textit{nOpp}). This variable measured the selected number of perspectives that oppose abortion. Similar to \textit{nCor}, it could take on seven different values (i.e., integers ranging from 0 to 6).\footnote{Here, we excluded topics that were used as attention checks. We do not compute the number of supporting perspectives selected due to symmetry.} 
\end{itemize}

\subsubsection*{\textbf{Individual differences}} We measured several variables that reflected individual differences among participants. These variables were later used to get a better idea of the sample as well as (in part) to answer \textbf{RQ2}.

\begin{itemize}
    \item \emph{Gender}. Selectable from multiple choices.
    \item \emph{Age}. Selectable by using a slider.
    \item \emph{Pre-existing stance}. Participants responded to the item ``\emph{In my opinion, abortion should be legal}'' by selecting the appropriate option from a 5-point Likert scale ranging from ``strongly disagree'' to ``strongly agree''.\footnote{Additionally, participants had the option to select an ``I don't know'' option. This option was also available for pre-existing knowledge.}
    \item \emph{Pre-existing knowledge}. Participants responded to the item ``\emph{I have good knowledge about the abortion debate}'' by selecting the appropriate option from a 5-point Likert scale ranging from ``strongly disagree'' to ``strongly agree''.
\end{itemize}

\subsubsection*{\textbf{Exploratory measurements}}

We used three additional items to measure the overall user experience with the task and to understand the possible potential a topic model has for a user. Participants could respond to each item by selecting the appropriate option from a 5-point Likert scale ranging from ``strongly disagree'' to ``strongly agree''. The results from these items were used for exploratory analyses.

\begin{itemize}
    \item \emph{Perceived usefulness}. To measure the general perceived usefulness of a model that can perform perspective discovery, participants responded to the item ``\emph{A model that can automatically show all viewpoints is useful to quickly understand a debate.}''
    \item \emph{Perceived awareness increase}. We measured whether participants experienced an increased awareness of the different perspectives related to \emph{abortion} by asking them to respond to the item ``\emph{I’m now better aware of the possible viewpoints than before.}''
    \item \emph{Confidence in task performance}. To measure participant's confidence in terms of whether the model helped them to make the right choices, participants responded to the item ``\emph{I’m confident that I’ve correctly assigned the viewpoints to the word groups}.''
\end{itemize}

\subsection{Procedure}
\label{section:method_procedure}

Our study consisted of an online task that we set up using the platform \emph{Qualtrics}.\footnote{\url{https://qualtrics.com/}} Before commencing with the study, participants had to agree to an informed consent. Both the study setup and the informed consent had been approved by the human research ethics committee at our institution before conducting this research. Participants then went through three subsequent steps:

\begin{figure}
    \centering
    \includegraphics[width=6cm]{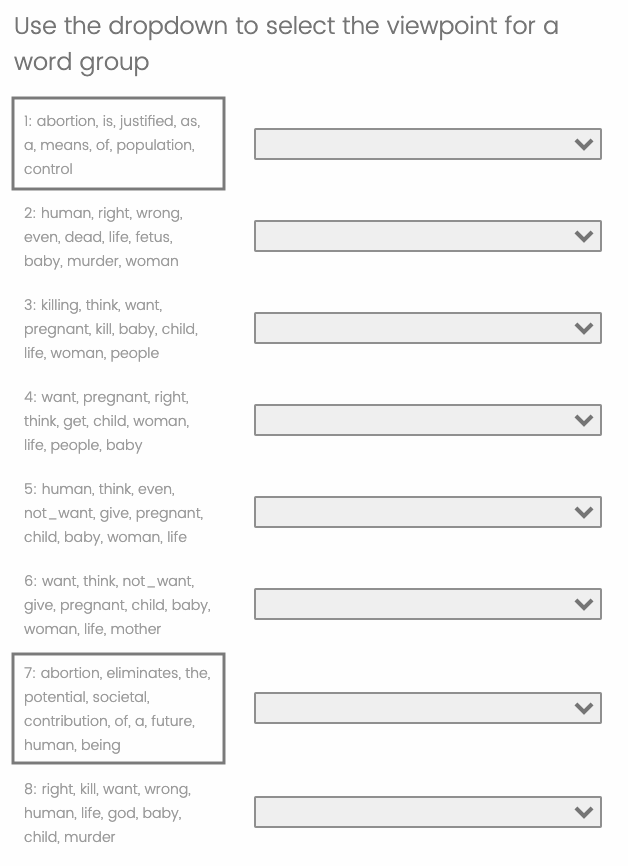}
    \caption{Screenshot of the main task. Word groups 1 and 7 (highlighted with a grey box) are the two honeypot topics.}
    \label{fig:mainTask}
\end{figure}

\paragraph{\textbf{Step 1}} Participants stated their age, gender, as well as pre-existing stance and knowledge related to \emph{abortion}.

\paragraph{\textbf{Step 2}} Participants did the main task. We randomly assigned each participant to one of the six models we aimed to test. After reading an introduction, participants were shown a list of 16 different perspectives. This list of 16 perspectives contained the six perspectives that were part of the corpus and ten other abortion perspectives taken from \emph{ProCon} (see Section \ref{data}).
Below the list of perspectives was the output of the model that participants had been assigned to. This output consisted of six ``topics'' that each were represented by a set of ten keywords (see Section \ref{sec:methodModels}). Additionally, we mixed two \emph{honeypot topics} into the output. Each of these honeypot topics was a set of keywords that matched one of the 16 perspectives word for word. 
Participants were instructed to match each set of keywords with one of the 16 abortion perspectives by selecting it from a drop-down menu (see Figure \ref{fig:mainTask}).

\paragraph{\textbf{Step 3}} We assessed participants' experience with the task. Specifically, we measured \emph{perceived model usefulness}, \emph{perceived awareness increase}, and \emph{confidence in task performance}. Additionally, participants were given the option to provide feedback using an open text field.

\subsection{Hypotheses}

Given our two research questions \textbf{RQ1} and \textbf{RQ2} as well as the operationalization and study procedure described above, we defined two hypotheses:

\vspace{.5cm}

\noindent \textbf{H1.} \textit{Users find more correct perspectives when being exposed to the output of a joint topic model compared to the output of a regular topic model or baseline.}

\vspace{.5cm}

\noindent \textbf{H2.} \textit{Users are more likely to identify sets of keywords as perspectives that are in line with their personal stance compared to perspectives that they do not agree with.}

\subsection{Statistical analyses}\label{section:statAnalysis}

Here, we describe the statistical analyses that we used to investigate \textbf{H1} and \textbf{H2}. All analyses were performed using either the open-source statistical software \emph{JASP}~\cite{JASP2020} or \emph{R}~\cite{R}. The \emph{JASP} file and \textit{R} code are openly available.\footnote{\url{https://osf.io/uns63/}}

\subsubsection*{\textbf{Investigating H1}}
We performed a one-way analysis of variance (ANOVA) with \emph{Model} as the between-subjects factor and \emph{nCor} as the dependent variable. This was to test the null hypothesis that there is no difference between models in terms of how many correct perspectives users were able to identify based on their output (i.e., the alternative hypothesis here was \textbf{H1}). 
Additionally, we checked the assumptions of normality and heterogeneity of variances using the Shapiro-Wilk and Levene's tests, respectively. In case the data did not meet the assumptions for the classical ANOVA, 
we would conduct a Kruskal-Wallis test as a non-parametric alternative.

In case we found a significant main effect of \emph{Model} on \emph{nCor}, we would perform posthoc tests to study which models specifically differ from each other. Because this series of posthoc tests would involve testing multiple (i.e., ${6\choose 2} = 15$) hypotheses, we would apply a Bonferroni correction to the traditional significance threshold of $0.05$ and therefore only regard $p$-values below $\frac{0.05}{15} = 0.003$ as significant.

\subsubsection*{\textbf{Investigating H2}}
We computed the Spearman rank correlation -- a non-parametric test for the correlation between two variables~\cite{spearman2008} -- between \emph{Pre-existing stance} and \emph{nOpp}. The null hypothesis in this test was that there is no correlation between these variables (i.e., the alternative hypothesis here was \textbf{H2}). Similar to other correlation coefficients, the Spearman rank correlation coefficient ranges from $-1$ to $1$. 

\subsection{Participants}

\begin{table}
\caption{Participant's pre-existing abortion stance.}
	\label{tab:stancedistribution}
	\centering
	{
		\begin{tabular}{lrrrr}
			\toprule
			\textbf{``In my opinion, abortion should be legal.''} & $\textbf{n}$ & \textbf{Percent} &\\
			\cmidrule[0.4pt]{1-4}
			Strongly disagree & 16 & 10.1 \\
			Somewhat disagree & 19 & 12.0\\
			Neutral & 16 & 10.1\\
			Somewhat agree & 26 & 16.5\\
			Strongly agree & 81 & 51.3\\
			Total & 158 & 100.0\\
			\bottomrule
		\end{tabular}
	}
\end{table}

To determine the required sample size for our study, we conducted a power analysis using the open-source software \emph{G*Power}~\cite{GPowerFaul2007}. Here, we specified an effect size $f = 0.3$, a significance threshold $\alpha = 0.05$, a statistical power of $0.8$, and a group size of $6$ (i.e., due to testing six different models). This resulted in a required sample size of at least $150$ participants. Based on a short pilot study we estimated that we would exclude about 10\% of participants due to failed honeypot checks. We thus recruited $170$ native English-speakers from the online participant pool \emph{Prolific}.\footnote{\url{https://prolific.co}} Here, we also applied an abortion-stance pre-screening offered by \emph{Prolific} to make the sample more balanced in terms of participant's personal attitude towards abortion (i.e., recruiting 135 ``pro-life'' and 135 ``pro-choice'' participants). After excluding some participants due to failing both honeypot checks, $158$ participants remained in the study.\footnote{To pass a honeypot check, participants had to allocate the right perspective to the honeypot topic that matched this perspective word for word (see Section \ref{section:method_procedure}.).}

Participants had a mean age of $33.34$ (ages ranged from $18$ to $64$). $49.4\%$ were male and $50.6\%$ female. Surprisingly, despite applying the abortion-specific pre-screening offered by \emph{Prolific} to approximate a 50/50 ratio in terms of participants who support/oppose abortion, participants in our sample turned out to largely support the legalization of abortion (see Table~\ref{tab:stancedistribution}). Most participants believed that they are familiar with the topic with $57.8\%$ responding with either ``strongly agree'' or ``somewhat agree''.

\section{Results}

In this section, we present the results of the hypothesis tests outlined in Section \ref{section:statAnalysis} and several exploratory findings.

\begin{figure}
    \centering
    \includegraphics[width=.8\linewidth]{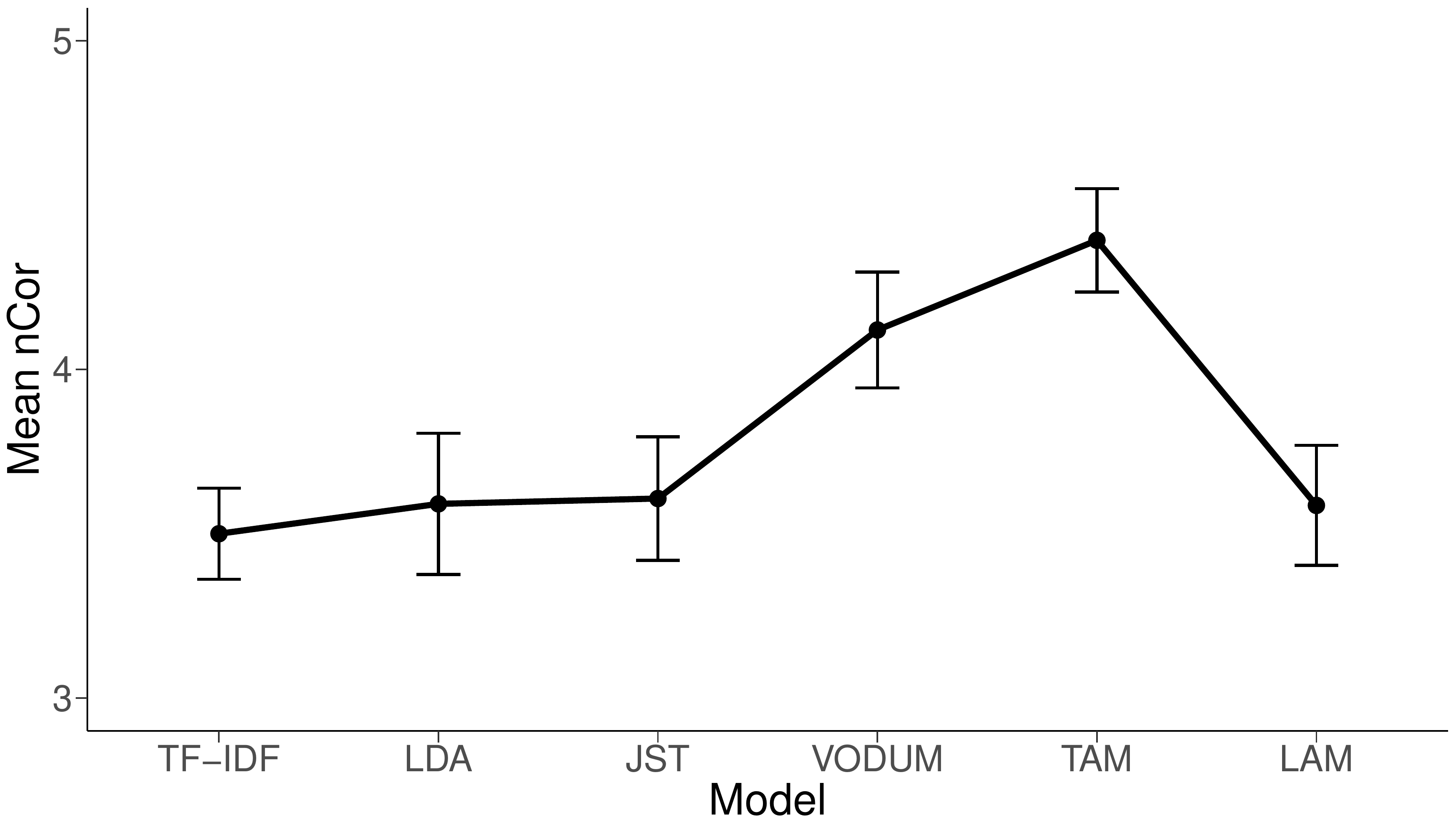}
    \caption{Mean \textit{nCor} (i.e., the mean number of correctly identified perspectives) per model. The error bars represent the standard error.}
    \label{fig:plot_perspectives}
\end{figure}

\subsection{H1: participants find more correct perspectives when using TAM}
We find that models differed in terms of how many of the six correct perspectives participants were able to identify. The ANOVA showed a significant main effect of \textit{Model} on \textit{nCor} ($F = 4.399, df = 5, p < 0.001, \eta^2 = 0.126$). Table \ref{tab:topicmodeldescriptive} and Figure \ref{fig:plot_perspectives} show the descriptive differences between the models with the highest mean \textit{nCor} for TAM ($4.39$). However, although the assumption of heterogeneity of variances held according to Levene's test ($F = 0.768, df = 5, p = 0.574$), the Shapiro-Wilk test suggested that the data were non-normal ($W = 0.905$, $p < 0.001$). We thus conducted a Kruskal-Wallis test as a non-parametric alternative to the classical ANOVA, which confirmed the results of the ANOVA ($X^2 = 20.611, df = 5, p < 0.001$). We therefore reject the null hypothesis that there is no difference between the models in terms of correctly identified perspectives.

\begin{table}[!h]
    \caption{Descriptive statistics of the user study. Here, $n$ refers to the number of participants, mean \textit{nCor} to the mean number of correctly identified perspectives per model (ranging from 0 to 6), and SE to the standard error.}
	\label{tab:topicmodeldescriptive}
	\centering
		\begin{tabular}{lrrr}
			\toprule
			\textbf{Model} & $\textbf{\textit{n}}$ & \textbf{Mean \textit{nCor}} & \textbf{SE} \\
			\midrule
			TF-IDF & 26 & 3.50 & 0.18 \\
			LDA & 22 & 3.59 & 0.19\\
			JST & 28 & 3.61 & 0.17\\
			VODUM & 25 & 4.12 & 0.18\\
			TAM & 28 & 4.39 & 0.17\\
			LAM & 29 & 3.59 & 0.17\\
			Total & 158 & & \\
			\bottomrule
		\end{tabular}
\end{table}

Due to the non-normality in our data, we conducted a series of non-parametric posthoc analyses (i.e., Mann-Whitney U tests) to study the individual differences between the models. The results show that only TAM led to significantly more correctly identified perspectives compared to the TF-IDF baseline model. 
Aside from that, the only significant difference we found was the one between TAM and LAM.


\subsection{H2: no evidence for user tendency to interpret model output in line with personal stance}

We did not find a significant correlation between \emph{pre-existing stance} and \emph{nOpp} ($\rho = 0.122$, $p = 0.163$). Based on these results, we cannot reject the null hypothesis that these two variables do not correlate. Our results thus do \emph{not} suggest that users are more likely to interpret the output of topic models in line with their personal stance.
  
\subsection{Exploratory results}

\begin{figure}
    \centering
    \includegraphics[width=.9\linewidth]{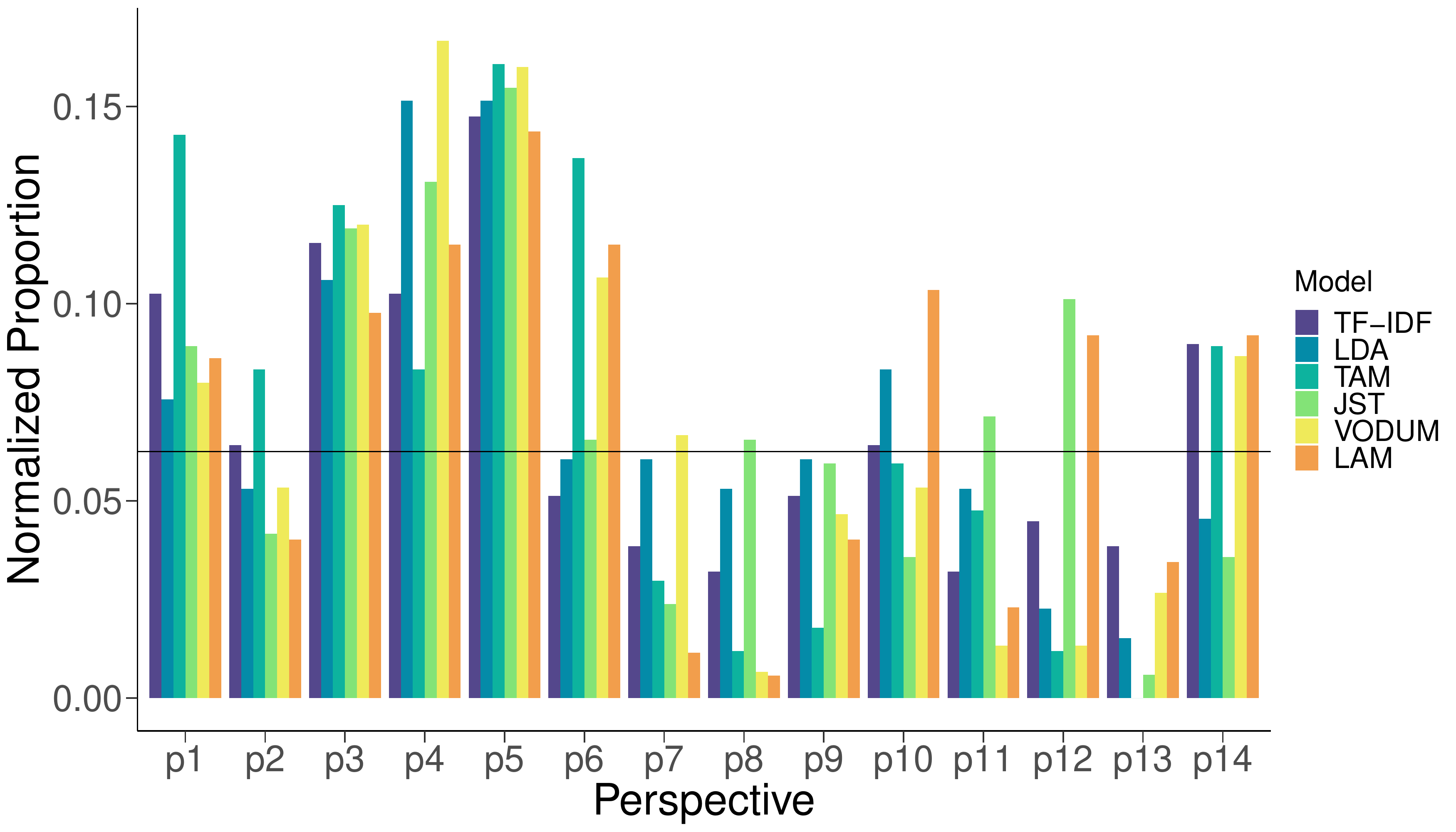}
    \caption{Normalized distribution of how often each available perspective was chosen (excluding the two honeypot checks). Whereas perspectives $p_1$ to $p_6$ were actually present in the corpus (see Table~\ref{tab:6perspectives}), the remaining perspectives were not. The red line is set to $\frac{1}{16} = 0.0625$.}
    \label{fig:distribution_sixperspectives}
\end{figure}

Figure~\ref{fig:distribution_sixperspectives} illustrates the normalized distribution of the chosen perspectives per topic model. It displays all perspectives that could be chosen in the task (excluding the two honeypot checks). The graph shows that some perspectives in the data (e.g., $p_5$) are more readily identified compared to other perspectives (e.g., $p_2$). Furthermore, we also see differences between the models that may help explain the results from the hypothesis tests. For instance, Figure~\ref{fig:distribution_sixperspectives} shows that, compared to the other models, TAM was a lot more successful in describing perspectives $p_1$, $p_2$, and $p_6$. TAM also did not lead people to false perspectives as much as other models did; for instance regarding $p_{10}$ and $p_{12}$.


\begin{table}
    \caption{Descriptive statistics on the exploratory measurements. Responses are from $5$-point Likert scales with $1$ = ``strongly disagree'' and $5$ = ``strongly agree''.}
    \label{tab:relationship_useful_confident}
    \centering
    \scalebox{0.775}{
    \begin{tabular}{lllllllll}
    \toprule
        \textbf{Question} & & \textbf{Overall} & \textbf{TFIDF} & \textbf{LDA} & \textbf{TAM} & \textbf{VODUM} & \textbf{JST} & \textbf{LAM}  \\
        \cmidrule[0.4pt]{1-9}
        Perceived & Mean & \textbf{$3.82$} & \textbf{$3.62$} & \textbf{$3.60$} & \textbf{$3.64$} & \textbf{$4.20$} & \textbf{$4.04$} & \textbf{$3.79$} \\
        Usefulness & Std & $1.06$ & $1.27$ & $1.0$ & $1.16$ & $0.76$ & $1.00$ & $1.01$\\
        \cmidrule[0.4pt]{1-9}
        Perspective & Mean & \textbf{$3.47$} & \textbf{$3.38$} & \textbf{$3.32$} & \textbf{$3.50$} & \textbf{$3.80$} & \textbf{$3.68$} & $3.17$\\
        awareness & Std & $1.13$ & $1.27$ & $1.09$ & $1.11$ & $1.00$ & $0.94$ & $1.31$\\
        \cmidrule[0.4pt]{1-9}
        Confidence & Mean & $2.83$ & $2.46$ & $2.68$ & $3.18$ & $2.72$ & $3.25$ & $2.62$\\
        & Std & $1.14$ & $1.24$ & $1.17$ & $1.12$ & $0.94$ & $1.08$ & $1.12$\\
        \bottomrule
    \end{tabular}
    }
\end{table}

Table~\ref{tab:relationship_useful_confident} shows descriptive statistics of the exploratory measurements as described in Section~\ref{section:operationalization}. Overall, participants reported high perceived usefulness of a model that can perform perspective discovery (mean = 3.82, sd = 1.06), indicating that they understood and approved of this method in general. Participants felt across models that their awareness of the different perspectives had increased (mean = 3.47, sd = 1.13 respectively), although this could be due to seeing the list of 16 possible perspective as opposed to a result of model performance. Confidence in task performance was not as high, with participants reporting moderate task performance confidence across models (mean = 2.83, sd = 1.14). This indicates that none of the models performed so well as to clearly communicate the different perspectives to users.


\section{Discussion}

We evaluated several joint topic models for the task of perspective discovery. Our results suggest that TAM can perform this task better than the TF-IDF baseline model. We find no evidence for a tendency of users towards interpreting model output in line with their personal stance.

Why did TAM perform better than other models? It seems that participants tried to find keywords in topics that explicitly appear in the perspective expression. For example, a topic containing the words \emph{God} and \emph{kill} is easily matched with perspective $p_5$ in our study (i.e., \emph{Abortion is the killing of a human being, which defies the word of God}). Whereas all models were able to distill this particular perspective quite well (see Figure \ref{fig:distribution_sixperspectives}), TAM also excelled at this task for other perspectives. Table~\ref{tab:computedtopics_tam} shows the TAM model output.

Outputting perspective-relevant keywords per topic seems to be a useful ingredient for a topic model that performs perspective discovery. Unlike the other joint topic models, TAM is \emph{designed} to distinguish common words appearing in any document and words being more topic-/perspective-specific. Models that use sentiment lexica to group words, such as JST and LAM, contained more sentiment words in their topic and were therefore less effective in discovering perspectives.

\begin{table}
\caption{The six topics computed by TAM.}
    \label{tab:computedtopics_tam}
    \centering
    \begin{tabular}{p{0.005\linewidth}p{0.875\linewidth}}
    \toprule
        $t_1$& woman, choice, body, fetus, control, pregnant, birth, baby, foetus, sex\\
        \cmidrule[0.4pt]{1-2}
        $t_2$ & fetus, human, brain, person, fetus\_not, cell, murder, alive, killing, egg\\
        \cmidrule[0.4pt]{1-2}
        $t_3$ & sex, woman, pregnant, parent, forced, child, want, child\_not, option, unwanted\\
    	\cmidrule[0.4pt]{1-2}
        $t_4$ & god, life, wrong, child, womb, baby, murder, killing, kill, creation\\
    	\cmidrule[0.4pt]{1-2}
        $t_5$ & want, woman, sex, not, responsibility, child, get, not\_want, pregnant, choice\\
    	\cmidrule[0.4pt]{1-2}
        $t_6$ & life, god, begin, baby, life\_begin, choice, choose, use, protection, responsibility\\
    \bottomrule
    \end{tabular}
\end{table}



\subsubsection*{Limitations and future work}

Our study is subject to several limitations. First, we created a data set containing debate forum entries with perspective annotations. This enabled us to curate a corpus of 600 documents that was balanced in terms of stance and perspectives. Such a scenario is unlikely to occur in real-world applications, where ``mainstream'' perspectives appear much more often than others. Second, despite our best efforts to control for it, our sample was not balanced in terms of pre-existing stance on the legalization of abortion: most participants turned out to support it. Third, we only evaluated one, highly politicized, commonly debated topic (i.e., abortion). It could be questioned whether the models we tested behave similarly on other, less divisive claims (e.g., \emph{zoos should exist} or \emph{social media is good for our society}). Fourth, although our results only show a difference between TAM and two other models, descriptive statistics suggest that there could be more subtle differences (see Figure \ref{fig:plot_perspectives}). If these differences truly exist, they could be discovered with a larger sample than the 158 participants we included in our study.

Future research could evaluate joint topic models for human-understandable perspective discovery using less balanced, more realistic data sets. We furthermore hope that our work inspires the creation of novel joint topic models that may outperform models such as TAM in perspective discovery. For instance, recent advancements in sentiment analysis such as word polarity disambiguation \cite{xia2015word} or predicting sentiment intensity \cite{wang2018using} could be incorporated to allow for a more fine-grained distinction between perspectives.

\section{Conclusion}

In this paper, we investigated whether joint topic models can help users distill perspectives from a corpus of opinionated documents. We find that a joint topic model such as TAM can indeed perform this task. Furthermore, we find no evidence for a tendency of users towards interpreting model output in line with their personal stance.

Our findings suggest that joint topic models have the potential to perform perspective discovery in a human-understandable way. If used in this way, they could find applications in many different areas, including policy-making or helping people overcome biases when participating in (online) debates. With the current trends towards global communication, such ways of structuring large corpora of opinionated documents seem ever more needed.



\bibliographystyle{IEEEtran}
\bibliography{sample-base}

\end{document}